\documentclass[manuscript,screen]{acmart}

\AtBeginDocument{%
  \providecommand\BibTeX{{%
    \normalfont B\kern-0.5em{\scshape i\kern-0.25em b}\kern-0.8em\TeX}}}

\setcopyright{acmcopyright}
\copyrightyear{2021}
\acmYear{2021}
\acmDOI{10.1145/1122445.1122456}

\usepackage{enumitem}
\begin{document}

\title{Ground-Truth, Whose Truth? - Examining the Challenges with Annotating Toxic Text Datasets}

\author{Kofi Arhin}
\email{arhink@rpi.edu}

\affiliation{%
  \institution{Lally School of Management, Rensselaer Polytechnic Institute}
  \streetaddress{}
  \city{Troy}
  \state{New York}
  \country{USA}
  \postcode{43017-6221}
}

\author{Ioana Baldini}
\affiliation{%
  \institution{IBM Research}
  \country{USA}}
\email{ioana@us.ibm.com}

\author{Dennis Wei}
\affiliation{%
  \institution{IBM Research}
  \country{USA}}
\email{dwei@us.ibm.com}

\author{Karthikeyan Natesan Ramamurthy}
\affiliation{%
  \institution{IBM Research}
  \country{USA}}
\email{knatesa@us.ibm.com}

\author{Moninder Singh}
\affiliation{%
  \institution{IBM Research}
  \country{USA}}
\email{moninder@us.ibm.com}

\renewcommand{\shortauthors}{Arhin et al.}

\begin{abstract}
  The use of machine learning (ML)-based language models (LMs) to monitor content online is on the rise. For toxic text identification, task-specific fine-tuning of these models are performed using datasets labeled by annotators who provide ground-truth labels in an effort to distinguish between offensive and normal content. These projects have led to the development, improvement, and expansion of large datasets over time, and have contributed immensely to research on natural language. Despite the achievements, existing evidence suggests that ML models built on these datasets do not always result in desirable outcomes. Therefore, using a design science research (DSR) approach, this study examines selected toxic text datasets with the goal of shedding light on some of the inherent issues and contributing to discussions on navigating these challenges for existing and future projects. To achieve the goal of the study, we re-annotate samples from three toxic text datasets and find that a multi-label approach to annotating toxic text samples can help to improve dataset quality. While this approach may not improve the traditional metric of inter-annotator agreement, it may better capture dependence on context and diversity in annotators. We discuss the implications of these results for both theory and practice.
\end{abstract}




\maketitle

\section{Introduction}
Language technologies have become important tools for content moderation on several platforms. Offensive content is either flagged or deleted completely. Some of these language technologies are trained on data that are labeled by annotators who provide ground-truth labels for the data samples. However, one of the main challenges with such datasets is the consistency and reliability of labels provided by annotators \cite{martin2021ground}. For example, text that is labeled as offensive in one context by an annotator, may be perceived differently by another. This may adversely impact the accuracy and fairness of models built using such datasets \cite{geva2019we}. Consistent with extant literature, we reason that the difficulty with annotator reliability and consistency is largely due to the fact that language is contextual, and its interpretation subjective \cite{wiebe2005annotating}. Thus, without context and clear guidelines, annotation tasks can create a lot of noise in benchmark datasets. 

To navigate the challenges with existing  datasets, several studies have suggested alternative approaches to annotation tasks and model development. For example, Matthew \textit{et al.} \cite{mathew2020hatexplain} posit that training models by highlighting the portion of a particular text that people use to distinguish offensive text from normal text can improve model performance. Also, Sap \textit{et al.} \cite{sap2019risk} show that priming annotators before annotation tasks can reduce their insensitivity to different dialects and the occurrence of bias in ground-truth labels. Similarly, Sap \textit{et al.} \cite{sap2019social} show how nudging annotators to provide additional information such as context inference, biased implications, and targets, among others, can help to improve the quality of crowdsourced datasets. However, Ball-Burack \textit{et al.} \cite{ball2021differential} find that solutions developed to tackle issues in one dataset may not necessarily be effective in resolving issues with out-of-sample datasets. In certain instances, annotator information may be required to improve model performance, highlighting the problem that labels may not be independent of annotators \cite{geva2019we}. These insights highlight the need for a deeper understanding of the issues with crowdsourced toxic text datasets \cite{geiger2021garbage}. Therefore, in this present study, we contribute to discussions on the difficulty with annotating toxic text datasets and highlight some recommendations to help improve annotator consistency and reliability. 

To help achieve the goal for this study, we adopt the design science research (DSR) \cite{hevner2004design} framework as a guide. The framework provides guidelines on developing innovative solutions to existing problems, especially where people and technology are concerned \cite{peffers2012design}, \cite{gregor2013positioning}. Using this framework as a guiding principle for problem identification and solution development, we find additional challenges to those that have already been highlighted in the extant literature. In examining three toxic text datasets that approach ground-truth labeling differently, we propose a multi-label approach to annotation. We use this multi-label approach to re-annotate three of the datasets, and find that 1) given different contexts, text samples can have different labels, 2) multiple labels for toxic text datasets can increase agreement with external ML annotators, but however, 3) this may not guarantee an improvement in inter-annotator agreement. We further discuss the implication of these results. 

The remainder of the study is organized as follows. In the next section, we briefly discuss some related studies. After that, we provide information on the selected datasets and methods. Next, we present the results from our analysis and discuss the implications for theory and practice. We conclude the study by highlighting the limitations of the study and opportunities for future research.

\section{Related Work}
\subsection{Ground-Truth in Toxic Text Classification}
For language and image classification tasks, ML models are built on datasets labeled by annotators. Through strategies such as majority voting, final labels are created and largely accepted as the ground-truth \cite{davani2021dealing}. For example, for language datasets, if annotators A and B label a text sample as "offensive", and annotator C labels it as "normal", the final label will be "offensive" based on majority voting. In some cases, annotators are compensated for their efforts \cite{snow2008cheap}. Therefore, it is important that the labels provided are accurate, and free of bias \cite{bender2018data}. However, this is not always the case as several challenges have been highlighted in existing studies. These challenges are not only limited to natural language. For instance, for image classification, Northcutt et al. \cite{northcutt2021pervasive} find significant errors in the labels for ImageNet, a popular image classification dataset. In one example mentioned in the study, a kayak was labeled as a dolphin. In this example, one can objectively say that there is an error with the kayak's label. For toxic text classification, the problem is more complex. It is not as straight-forward to claim that a text sample has been given the wrong label. One of the reasons for this is the highly contextual nature of language. For instance, a phrase or sentence that is perceived as offensive in one context may be labeled as normal in another context. On the contrary, we can argue for example that a kayak will remain a kayak under all contexts. In addition to context, the background and values of the annotator play a significant role in the final label. As such, bound by the same context, a text sample can be labeled differently by different annotators. Hence, in some instances, ML models built on toxic text datsets may require annotator information to boost prediction accuracy \cite{geva2019we}. These challenges may adversely impact model performance metrics and point to the need to review the process of labeling data, especially for natural language datasets.

To a large extent, the challenges identified in labeling toxic text datasets fall under the broad categories of inter-annotator agreement and annotator consistency. Inter-annotator agreement refers to the extent to which annotators provide the same label for a particular text sample \cite{artstein2017inter}. On the other hand, annotator consistency refers to the extent to which a single annotator provides the same label for similar text samples \cite{ishita2020using}. Therefore, achieving high agreement and consistency rates in a dataset may have positive implications for model performance. In existing studies, metrics such as the Krippendorff's alpha \cite{hayes2007answering} and Choen's kappa \cite{cohen1960coefficient} have been used as indicators of annotated data quality and reliability. In addition to these challenges, at the training and predictions phases of toxic text classification, metrics such as accuracy and fairness are important in reviewing ML model performance. Especially for toxic text classification, in addition high prediction accuracy, researchers are concerned with ensuring that language constructs that are associated with minority groups are not falsely classified \cite{ball2021differential}. Largely, strategies to resolve some of the issues are implemented at the data processing, modeling and post-modeling stages in the ML pipeline. We are yet to fully understand how strategies can be implemented at the data collection stage to address some of the identified challenges. Thus, to contribute to efforts in this regard, we adopt a design science research approach \cite{hevner2004design} to examine selected toxic text datasets, discuss some identified challenges, and share insights from our proposed approach.

\subsection{Design Science Research (DSR)}
The DSR framework provides guidelines on creating innovative solutions to address problems, especially where people and technology are concerned \cite{hevner2004design}. The stages outlined by the proponents of the DSR framework include 1) problem identification, 2) objectives of a solution, 3) design and development, 4) demonstration, 5) evaluation, and 6) communication \cite{peffers2012design}. The problem identification stage is mainly concerned assessing the issues related to the problem at hand. The objectives phase is meant to provide directions for addressing the challenges. The design and development, demonstration, and evaluation stages focus on executing the proposed solutions and evaluating the extent to which the objectives are achieved. Beyond all this, it is important to effectively communicate the entire project to interested parties.

In addition to the DSR stages, some studies establish that design research should lead to two outcomes namely, a process, and/or an artifact \cite{gregor2013positioning}. In this study, we introduce a process for annotating toxic-text datasets as a strategy to address key issues that affect dataset quality.  Therefore, an effective design science solution requires a good understanding of the existing problem \cite{storey2017using}. One of the ways to achieve this is by making use of case studies \cite{aken2004management}. Therefore, using these as guiding principles, we examine three toxic text datasets, and propose solutions to addressing some of the challenges. The overarching objective therefore is to enumerate the issues identified in the datasets, provide insights on their impact, and assess the extent to which the solutions we propose can address some of the challenges identified in these case studies. 

\section{Data and Methods}
\subsection{Data}
The datasets selected for the study are the HateXplain \cite{mathew2020hatexplain}, Social Bias Inference Corpus (SBIC) \cite{sap2019social}, and the Jigsaw\footnote{https://www.kaggle.com/c/jigsaw-toxic-comment-classification-challenge} datasets. Our selection of these three datasets is founded on the basis that they address a similar problem (toxic text), yet they are diverse in how the annotations were collected. For instance, HateXplain has exactly three annotators for all text samples while the SBIC and Jigsaw datasets do not have a fixed number of annotators for samples.  The SBIC dataset has additional background information on annotators (i.e., data statements \cite{bender2018data}). Further, while the HateXplain and SBIC sets use majority voting to determine the final label, the Jigsaw data uses a continuous final label (i.e, toxicity) that represents the proportion of annotators who labeled a particular sample as toxic. For example, if 4 out of 5 annotators label text sample \textit{A} as offensive, Jigsaw's final label will be 0.8 toxicity, while the HateXplain and SBIC datasets will have \textit{offensive} or \textit{hatespeech} as the final label. Table 1 provides a summary of the number of annotators, text samples, and the distribution of final labels. In the last column for Table 1, we include our definition for offensive text for each dataset for the purpose of this study. Appendix A shows text samples from the three datasets.

\begin{table}[h]
  \caption{Dataset Summary}
  \label{data-summary}
  \centering
  \begin{tabular}{p{1.5cm}p{2cm}p{2cm}p{2cm}p{1.5cm}p{3cm}}
    \toprule
    Dataset &  Unique \newline Text Samples &   Offensive \newline Text &  Normal \newline Text & Total \newline Annotators & Offensive \newline Text Definition \\
    \midrule
    HateXplain  & 20,148 & 12,334 (61\%) & 7,814 (39\%) & 253 & Both \textit{hatespeech} and \textit{offensive} text  \\
    SBIC     & 45,318 & 25,073 (55\%) & 19,401 (45\%)  & 307 & Samples given the labels 1 and 0.5   \\
    Jigsaw     & 1,804,874 & 120,084 (7\%) & 1,684,790 (93\%) & 8,899 & Samples with toxicity 0.5 and above\\
    \bottomrule
  \end{tabular}
\end{table}

\subsection{Sampling}
To help with identifying issues, we re-annotated one hundred (100) randomly selected samples from each of the datasets. In Table 2 below, we show the distribution of the selected samples for each dataset. The samples were found to match the label distribution of offensive versus normal text samples in the original datasets. With the exception of the SBIC dataset, each of the samples was annotated by all five (5) authors of this study.

\begin{table}[h]
  \caption{Sampled Data}
  \label{samples-summary}
  \centering
  \begin{tabular}{p{1.5cm}p{2cm}p{2cm}p{2cm}p{1.5cm}p{3cm}}
    \toprule
    Dataset &  Unique \newline Text Samples &   Offensive \newline Text &  Normal \newline Text & Total \newline Annotators & Offensive \newline Text Definition \\
    \midrule
    HateXplain  & 100 & 61 & 39 & 5 & Both \textit{hatespeech} and \textit{offensive} text  \\
    SBIC     & 100 & 60 & 40  & 3 & Samples given the labels 1 and 0.5   \\
    Jigsaw     & 100 & 9 & 91 & 5 & Samples with toxicity 0.5 and above\\
    \bottomrule
  \end{tabular}
\end{table}

\subsection{Team Annotation Task Design}
Earlier, we established that clear instructions are necessary to serve as a guide to annotators on how to handle annotation tasks.  Therefore, similar to Guest et al. \cite{guest2021expert}, we develop new guidelines to guide annotators for this task. In addition to the guidelines, we reason that it is important to allow annotators to skip text samples that are difficult to categorize due to the occurrence of contextless samples. Therefore providing a third label such as \textit{undecided} can help researchers identify problematic or difficult samples in the data. Appendix B contains the guidelines provided for the annotation task.

To address the issues of context, consistency and agreement, we propose three context-based label columns for re-annotating the three datasets. Our justification for this proposal is founded on our examination of existing datasets which supported the assertion that language is highly contextual. Hence, providing multiple contexts for labels may help to improve consistency and agreement between annotators. The label columns for the team annotation task are: \textit{strict label}, \textit{relaxed label}, and \textit{inferred group label}. For the \textit{strict label}, we asked annotators to consider the task as a bag-of-words approach where the appearance of certain words in a text makes the entire text either offensive or normal. For the \textit{relaxed label}, we asked annotators to consider any instance where an offensive text sample, under the strict label could be considered as normal. For the \textit{inferred group label} we asked annotators to consider whether an offensive text sample can be considered normal if it were uttered by a member of the identity group in the text.

In addition, we conducted a separate exercise to assess the extent to which the proposed multi-label columns can make annotators consistent. Therefore, we divided the team into two. Three team members labeled the same samples twice, at least one week apart using the multiple labeling scheme, while two of the team members used a single column for labeling the datasets. For this task, we used only the samples from the SBIC dataset. 

Due to the limited number of annotators and samples, we present our results as propositions that can be tested further and validated statistically. We understand that additional studies with a larger sample pool can help make stronger and more generalizable claims.

\subsection{External Annotators}
Since we re-annotated a limited number of samples from the three datasets (i.e., one hundred samples from each), we used ML-based tools as external annotators to validate the labels provided since we did not have enough samples to split the data training and prediction by an ML model. We therefore used Detoxify \cite{Detoxify} and Perspective API\footnote{https://www.perspectiveapi.com/} as external annotators. In doing this, we were able to compare the final labels from the re-annotation task with the original labels provided in the datasets. The toxic text classification frameworks for Detoxify were built on two different datasets, leading to the development of two different packages; Detoxify Original (Dtx(Og)) built on Wikipedia comments and Detoxify Unbiased (Dtx(Unb)) built using the Jigsaw data. The Perspective API (PsAI) framework was built in a collaborative research effort between Jigsaw and Google.

In addition, Perspective API has four distinct categories for offensive text, namely toxic, obscene, insulting and threatening. For the purpose of this study, we deem any sample that scored 0.5 and above as offensive in these four categories. Detoxify also has scores for different categories, namely toxicity, severe toxicity, obscene, threat, insult, identity attack. Again, for the purpose of this study, we focus on the scores for toxicity and label any sample that scores 0.5 and above as offensive. These two popular toxic classification frameworks provide a good foundation for evaluating the final labels. In the sections that follow, we present our findings.

\section{Findings}

\subsection{Problem Identification}

\begin{table}[htbp]
  \caption{Some Identified Challenges in the HateXplain, SBIC, and Jigsaw Datasets}
  \label{challenges}
  \centering
  \begin{tabular}{p{7cm}p{7cm}}
    \toprule
    Problem Identified &   Implications  \\
    \midrule 
    \textbf{Annotator Influence:} \newline
    In HateXplain, one annotator contributed to the final label for 5,730 samples. 5,730 samples means a single annotator contributed to about 28\% of the final labels, about 3,000 samples and 18\% points more than the second-ranked annotator in the HateXplain set.  
    \newline
    \newline
    In SBIC, a majority (i.e., 80\%) of the annotators identified as white. We examined whether people in the minority groups were more likely to oppose the final label and found for example 17.4\% of annotators who identified as black, opposed the final label at least once, while this figure was 12\% for the white annotators. In Appendix C, we show the opposition rate by race.
    
    &
    \medskip
    Since annotator behavior can be reflected in prediction outcomes \cite{geva2019we}, \cite{bender2018data}, this is problematic because the annotator highly influences the outcome (final label) of many samples. This can lead to one annotator's behavior being further amplified by a ML model.
    \newline
    \newline
    Low diversity in annotators selected for a labeling task can result in the silencing of minority voices. Since bias is one of the key concerns for recent advances in toxic text classification, it is important to understand how low diversity can impact predictions.
    \\    
    \textbf{Annotation (Im)Balance:} \newline
    In Jigsaw and SBIC, we found instances where a large pool of annotators contributed to the final label for some text samples while others received labels from relatively fewer annotators. For example, the minimum annotators per text for the Jigsaw data is 3 annotators, while the maximum annotators per text is 4,936 annotators. As such, similar samples may have different toxicity rates due largely to very different numbers of annotations. In Appendix D, we show the distribution of sample annotaions for the SBIC and Jigsaw datasets.
    
    & 
    \medskip
     This is a problem, especially for the Jigsaw dataset, because the final label is the proportion of annotators who label a sample as offensive. Consider text sample \textit{B} labelled by 3 annotators and \textit{C} labeled by 1000 annotators. For sample \textit{B}, if even one annotator labels the text as offensive, toxicity will equal 0.33. However, for \textit{C}, if 100 annotators label the sample as offensive, toxicity will equal 0.10, suggesting that sample \textit{B} is more toxic than \textit{C} which might not be the case.
    \\
    \textbf{Inconsistent Labels:} \newline
    We found instances where annotators provided different labels for similar text. For example, in HateXplain, \textit{has stupid rich h*e} was labeled as \textit{normal} while \textit{...b**** a** back to the east} was labeled  \textit{offensive} by the same annotator. Appendix E shows another instance of this in the SBIC dataset.
    
    & 
    \medskip
    Inconsistency has been one of the major concerns for building good models using toxic text datasets. It creates noise in the data which ultimately leads to poor model performance \cite{gordon2021disagreement}, \cite{ishita2020using}.
    \\
    % \smallskip
    \textbf{Contextless Samples:} \newline
    Some text samples were difficult to place in specific contexts. This made it difficult to know which labels to assign to them. However, annotators provided labels for these sample. For example, the sample \textit{"why Arabs lose wars"} from HateXplain is difficult to place in a specific context, making it difficult to categorize. In Appendix F, we show more examples of contextless samples.
    & 
    \medskip
    Contextless samples also have the tendency to lead to noisy labels because of the high level of uncertainty. This can also lead to large disagreement rates between annotators.
    \\
    \bottomrule
  \end{tabular}
\end{table}

Table 3 above provides a summary of the problems identified in the datasets. The identified issues are grouped into four main categories, namely Annotator Influence, Annotator (Im)Balance, Inconsistent Labels, Contextless Samples.

\subsection{Distribution of Labels}

Using the proposed multi-label approach from the guidelines in Appendix B, we re-annotated one hundred samples each from all three datasets. The samples were annotated separately by the five authors and the final labels were determined by majority vote. Table 4 shows the distribution of the final labels from the original datasets (Og.) and three columns; Strict (St.), Relaxed (Rel.), and Inferred Group (Ig.) The distributions show that given different contexts (i.e. Str., Rel., and Ig.)), the final labels can change for a particular text sample. We found that for each of the proposed three columns, the final labels did not align perfectly with the original benchmarks, emphasizing our point for contextual labeling. We also found that if given the option, annotators may be undecided on some text samples.

\begin{table}[htbp]
   \caption{Label Distribution from Team Annotation Task}
   \label{label-distribution-team}
   \centering
   \begin{tabular}{l|cccc|cccc|cccc}
     \toprule
     & \multicolumn{4}{c|}{HateXplain} &  \multicolumn{4}{c|}{Jigsaw} & \multicolumn{4}{c}{SBIC}                \\
     Label     & Og. & Str. & Rel. & Ig. & Og. & Str. & Rel. & Ig. & Og. & Str. & Rel. & Ig. \\
     \midrule
     Normal & 61  & 68 & 61 & 59 & 91  & 96 & 93 & 93 & 40  & 59 & 46 & 51  \\
     Offensive & 39 & 32  & 39 & 40 & 9  & 4 & 6 & 6  & 60  & 30 & 51 & 46   \\
     Undecided & -  & -  & - & 1 & -  & - & 1 & 1  & -  & 1 & 3 & 3  \\
     \bottomrule
   \end{tabular}
    \small
    \\
    \bigskip
        Og.: original dataset label, Str.: strict label, Rel.: relaxed label, Ig.: inferred-group label
\end{table}

\subsection{Label Agreement}
In the previous section, we showed that different contexts can lead to different labels for text samples. In Table 5 below, we show the extent to which the original labels and final labels agreed. We find that to a large extent, our relaxed label and inferred group labels have a relatively higher agreement with each other. However, our strict label shows a much higher rate of disagreement with the original and relaxed labels for all datasets. Out of the three datasets, the Jigsaw data shows a much higher agreement rate with our new labels. It is worth noting that the Jigsaw data contains only a small percentage of offensive text compared to the other two datasets. We can therefore infer that offensive text is perhaps more difficult to label.

\begin{table}[htbp]
  \caption{Label Agreement from Team Annotation Task}
  \label{agreement-rate}
  \centering
  \begin{tabular}{lrrrr}
    \toprule
    Column &  HateXplain &   Jigsaw & SBIC \\
    \midrule
    Strict-Relaxed & 0.87 & 0.95 & 0.59\\
    Strict-InGroup    & 0.85 & 0.95 & 0.61 \\
    Strict-Original     & 0.77 & 0.91 & 0.54\\
    Relaxed-InGroup     & 0.97 & 1.00 & 0.96\\
    Relaxed-Original     & 0.80 & 0.92 & 0.86\\
    InGroup-Original     & 0.78 & 0.92 & 0.84\\
    \bottomrule
  \end{tabular}
\end{table}

\subsection{Label Validation}
As mentioned in the methods section, to complement the label agreement results, we used two ML tools (external annotators) to assess the final labels for the original and new labels, namely Perspective API\footnote{https://www.perspectiveapi.com/} and Detoxify \cite{Detoxify}. Our aim for doing this is provide a proxy for label quality due to the limited samples we re-annotated. The results in Table 6 show that the new multi-labels have relatively higher agreement rates with Perspective API and Detoxify for all three datasets compared to the original labels. For the Jigsaw dataset, it was not surprising to see that original label had a higher agreement with Dextoxify(Unbiased) and Perspective API because both frameworks were developed using the Jigsaw dataset. . These results suggest to some extent that, providing multiple labels for different contexts can improve model performance, including out-of-sample predictions. We attribute the comparatively high agreement rate between the original Jigsaw labels and the Dtx(Unb) labels to the fact that framework was built using the Jigsaw dataset. 

\begin{table}[htbp]
   \caption{Label Agreement with External Annotators}
   \label{label-agg}
   \centering
   \begin{tabular}{l|cccc|cccc|cccc}
     \toprule
     & \multicolumn{4}{c|}{HateXplain} &  \multicolumn{4}{c|}{Jigsaw} & \multicolumn{4}{c}{SBIC}                \\
     
     Label     & Og. & Dtx & Dtx(Unb). & PsAI.& Og. & Dtx & Dtx(Unb). & PsAI & Og. & Dtx & Dtx(Unb). & PsAI \\
     \midrule
     Strict &  0.77 & 0.74 & 0.75 & 0.75 & 0.91 & 0.97 & 0.94 & 0.91 & 0.54 & 0.83 & 0.83 & 0.75\\
     Relaxed & 0.80 & 0.65 & 0.66 & 0.68 &  0.92 & 0.96 & 0.93 & 0.92 & 0.86 & 0.58 & 0.56 & 0.58\\
     InGroup & 0.78 & 0.63 & 0.66 & 0.66 & 0.92 & 0.96 & 0.93 & 0.92 & 0.84 & 0.58 & 0.56 & 0.58\\
     Original & - & 0.61 & 0.60 & 0.62 & - & 0.92 & 0.97 & 0.96 & - & 0.56 & 0.54 & 0.60\\
     \bottomrule
   \end{tabular}
   \small
    \\
    \bigskip
    Og.: Original dataset label, Dtx.: Detoxify original, Dtx(Unb).: Detoxify unbiased, PsAI.: Perspectve API
\end{table}

\subsection{Inter-Annotator Agreement}
According to extant literature, inter-annotator agreement is an important component of natural language datasets. For the purpose of this study, we define inter-annotator agreement as the extent to which text samples received a unanimous (whether offensive or normal) vote from annotators. In using a multi-label approach, we expected a high agreement rate between annotators. This is because we believed context-based columns could provide more clarity to annotators. However, this was not the case. In Table 7 below, we show that in some cases the new multi-label columns performed relatively poorer with regards to the overall inter-annotator agreement. Perhaps, this poor performance supports our argument that toxic text samples are difficult to label. 

\begin{table}[htbp]
   \caption{Rate of Agreement between Annotators}
   \label{label-interannotator-agreement}
   \centering
   \begin{tabular}{l|cccc|cccc|cccc}
     \toprule
     & \multicolumn{4}{c|}{HateXplain} &  \multicolumn{4}{c|}{Jigsaw} & \multicolumn{4}{c}{SBIC}                \\
     
     Label     & Og. & Str. & Rel. & Ig. & Og. & Str. & Rel. & Ig. & Og. & Str. & Rel. & Ig. \\
     \midrule
     Normal & 0.74  & 0.46 & 0.55 & 0.60 & 0.77  & 0.84 & 0.68 & 0.69 & 0.70  & 0.78 & 0.78 & 0.76  \\
     Offensive & 0.42 & 0.70  & 0.61 & 0.60 & 0.11 & 0 & 0 & 0 & 0.78  & 0.57 & 0.55 & 0.61   \\
     Undecided & -  & -  & - & 0 & -  & - & 0 & 0 & -  & 1.00 & 0 & 0 \\
     Overall & 0.62 & 0.61  & 0.58 & 0.58 & 0.71 & 0.82 & 0.64 & 0.65 & 0.75  & 0.64 & 0.64 & 0.67  \\
     \bottomrule
   \end{tabular}
   \small
    \\
    \bigskip 
    Og.: original dataset label, Str.: strict label, Rel.: relaxed label, Ig.: inferred-group label
\end{table}

\subsection{Annotator Consistency (Intra-Annotator Agreement)}
In addition to inter-annotator agreement, another important component of ground-truth labels is annotator consistency. As outlined earlier, consistency refers to the extent to which similar samples are given similar labels by a single annotator. As highlighted in the methods section, we used only the samples from the SBIC dataset, dividing the team members into two groups with three and two team members. The results from the consistency task showed that, the team members who used the single column had an average of 91\% consistency while the team members who used the multi-label columns had an average of 86\%, 89\%, and 89\% for the strict, relaxed and inferred-group labels respectively. Again, while we are careful not to draw any statistical conclusions from this, we found that for the multi-label team, the highest level of consistency for the multi-label columns were 99\%, 97\%, and 97\% for the strict, relaxed and inferred-group labels respectively. However, the highest level of consistency for the single column team was 92\%. 

In the section that follows, we discuss the implications of our results for both theory and practice.

\section{Discussion}
This study shares insights on why labeling toxic text datasets is a difficult task. By examining three existing datasets, we highlighted challenges that reinforce and complement
existing issues pointed out in extant literature. The goal was not to show how poor the existing datasets are, but rather contribute to discussions on how to address challenges for existing and future annotation tasks. The results from the problem identification and re-annotation tasks point to five main insights. 

First, the fact that language is highly contextual signifies that a multi-label approach may be more beneficial for ML models and classification tasks. Our re-annotation task showed that the same text sample can have different labels given different contexts. To provide support for this first insight, the results in Table 6 showed a relatively higher agreement rate between the multi-label approach and external annotators. This suggests that using multiple labels can lead to positive outcomes such as less noisy datasets and improved metrics for out-of-sample predictions. To a large extent, ML models built using toxic text datasets are trained with a single column as the dependent (target) variable. Since our approach creates three columns, developers can first  consider the context of a specific task, and then use the most suitable column in a multi-label scenario. Thus, a multi-label approach can afford developers the flexibility to switch between contexts. One interesting possibility for taking this a step further is to use \emph{multi-label classification} \cite{tsoumakas2007multi-label}, which builds a single ML model to predict all labels simultaneously and can exploit relationships between the labels. %\DW{Perhaps we could explicitly mention the area of multi-label classification in ML?} 
In the absence of multiple labels, we propose that ML models can be trained to first understand the context of text samples before making predictions. 

Second, one must reckon with the fact that disagreement in a general sense increases as the number of annotators increases. Taking the rate of unanimous agreement as a simple measure, for one hundred samples from the HateXplain dataset, the agreement rate at two annotators was 81\% but gradually decreased to 56\% when three additional annotators were added. Looking at the inter-annotator agreement rate separately for offensive and normal text samples, our expectation that offensive text would be more difficult to label seem to hold for the original labels in the datasets. For the team annotation exercise, while this was true for both the SBIC and Jigsaw datasets, the HateXplain labels showed a different pattern where the team had a higher agreement rate for offensive text. Understanding the relationship between text categories and agreement rates can help to tailor solutions to identified challenges. Hence, we believe this issue is worth exploring further to improve our understanding of the phenomenon.

Third, while the results for the inter-annotator agreement task was not as expected across all datasets, it is worth emphasizing that this may not signal poor dataset labels. Rather, contrary to some existing studies, we believe that annotator disagreement could be an indication of annotator diversity, which should be a desirable attribute. Annotator diversity ensures that members of a minority group are not adversely impacted by prediction tasks as has been discovered by some studies. For instance, Ball-Burack et al.~\cite{ball2021differential} found that text samples high in language from the African-American English dialect are more likely to be labeled as offensive. Our discussions at the problem identification stage of this study also show that annotators who belonged to minority groups were more likely to oppose the final label compared to annotators in the majority group. Hence, in agreement with Davani et al. \cite{davani2021dealing} we believe it is important to understand why annotators disagree rather than trying to achieve high agreement rates which may lead to biased outcomes.

Fourth, %we found that a multi-label approach, if implemented well, can help annotators achieve a good level of consistency. 
while the average consistency scores from our preliminary re-annotation experiment %for consistency 
did not of course provide %the expected 
conclusive results, we are intrigued by the observation that high levels of consistency are possible with the multi-label approach. %comparing the highest level of consistency achieved by the annotators from the two groups showed that the multi-label approach can help to improve the consistency of annotators. \DW{flagging for later to think about phrasing more carefully, to not oversell one data point} 
We believe that intra-annotator consistency is a significant benchmark for dataset quality, especially for toxic text datasets. As such, there should be additional efforts both to answer our particular question about single-label versus multi-label as well as to support research in this area more generally. 

Fifth, for the SBIC dataset, we found that while some offensive text samples were easy to identify, a significant number were more difficult to label. This is mainly because the difficult ones did not have words that were toxic. To a large extent, toxicity was implied and not as glaring as in the other samples. Take for example the text samples from Table 8 below which are extracts from the SBIC dataset, labeled as offensive by both the team (under the strict label) and the original labels: 

\begin{table}[htbp]
  \caption{Implicit Offensiveness}
  \label{imp-offensive}
  \centering
  \begin{tabular}{p{10cm}}
    \toprule
    Sample Text from SBIC \\
    \midrule
    \textbf{Sample A} \\
    \textit{"i really HATE a ""come over"" ass n***** like wtf every time we communicate dont tell me to come over tf take me out or something f***"}\\
    \\
    \textbf{Sample B} \\
    \textit{"what 's the difference between an <number> yr old girl and a washing machine? when you dump a load in the washer , it will not follow you around for two months"}\\
    \\
    \bottomrule
  \end{tabular}
\end{table}

While text sample A is easy to classify as offensive focusing on keywords as directed in our guidelines, text sample B requires a deeper understanding of what is being implied. This could also have adverse implications for ML models when there are no keywords or phrases to focus on. In addition to the difficulty of identifying the toxicity, the label also heavily depends on annotator characteristics such as demographics, values, emotion, humour, among others. %\DW{Should we take this opportunity to comment further on how sample A could be construed as not offensive? i.e.~it could be someone complaining about their partner to their close African-American friends} 
Difficult samples such as these accounted for inconsistent labels for the most part. 

\section{Limitations and Future Work}
While this study provides useful insights for toxic text datasets, there are some notable limitations. One of the main limitations of this study is the fact that although we reviewed three toxic text datasets with a large number of samples, we only annotated one hundred (100) samples from each of them. While a hundred samples may not be enough to provide an exhaustive list of insights, we believe that we have revealed important insights that complement the findings in the existing literature.

Furthermore, the study focuses only on three datasets as case studies. Three out of the many toxic text datasets may not be representative enough. Since there are several toxic text datasets with unique attributes, our findings may not be applicable to all of them. However, our propositions and findings may cut across many datasets because of the broad issues of annotator agreement and consistency. 

Our preliminary experiment to understand how the multi-label and single-label columns both impact consistency was limited to 5 annotators. %The constraint of recruiting annotators prevented using a large number of people and so we had to default to the authors. 
For future research projects, we hope to recruit additional annotators to help us make adequate statistical inferences. 

For future studies, we hope to address these challenges to provide additional insights to complement those discussed in this study. In spite of these limitations, we believe that the insights shared here can provide useful implications for theory and practice and can result in new directions/hypotheses that can be tested by future research studies.

\section{Conclusion}
In this study, we highlight the need to pay attention to the issue of context in existing and future toxic text datasets for NLP tasks. We enumerate the challenges in some selected datasets and add to conversations pertaining to addressing them. We find that while language is difficult to annotate, using multiple columns labels can help to reduce some of the identified challenges. We found that a multi-label approach may also has positive implications for reducing noise in datasets which can also improve out-of-sample predictions.

\bibliography{references}
\bibliographystyle{abbrv}

\clearpage
\appendix
Appendices

\section{Text Samples from Datasets}
\begin{table}[h]
  \caption{Text Samples from Datasets}
  \label{text-samples}
  \centering
  \begin{tabular}{p{2cm}p{8cm}p{2cm}}
    \toprule
    Dataset &  Sample Text & Final Label\\
    \midrule
    HateXplain & \textit{"perhaps a cause of such senseless violence we see in our streets"} & Normal\\
    \\
    &  \textit{"<user> you sound like a huge faggot fuck off kike lover"} & Offensive\\
    \\
    SBIC & \textit{"Even if I get braces I'll be in my mid 20's by the time they're off"} & Normal\\
    \\
    & \textit{"I just threw a string of swears at the tv, the likes of which no one has ever ball sackin' goddamn fucking ass twat seen."} & Offensive\\
    \\
    Jigsaw & \textit{"It's great if you can afford it. When we retired, we sold and bought an antebellum house in the south. Life is good!"} & Normal\\
    \\
    & \textit{"Besides Planned Parent hoodlums need the body parts. (Gross but true)"} & Offensive
    \\
    \bottomrule
  \end{tabular}
\end{table}

\section{Annotation Guidelines}
Instruction for Annotators:
It is recommended that annotators read through the assigned batch of examples to calibrate before annotating.

\textbf{1. Label Types}

The label types for the sample datasets are offensive, normal or undecided.

\textit{Normal (0):} 
Non-abusive text

\textit{Offensive (1)}: 
Abusive or toxic words, especially directed towards a protected group. This can include abbreviations of swear words (eg. "af", "lmao"). A text can also be classified as offensive even if there are no derogatory terms, insults, express violence, etc. used, but offensiveness is implied. 

\textit{Undecided (2)}: 
When the annotator is not sure which category an individual sample belongs to. This sometimes includes contextless samples, and samples with borderline words that make it difficult to classify the text.

\textbf{2. Label Columns}

These are the label columns required to be completed by the annotator. The rationale for these columns were initially discussed at our team meeting on Wednesday June 16, 2021 and subsequently modified. They are 'strict label', 'relaxed label', and 'inferred group label'.

\textit{a. Strict label}: 
Anytime a word that you think is offensive is spotted (based on your mental list of words or phrases), the whole sample will be labeled offensive. If there are no offensive words, then it is normal. 

\textit{b. Relaxed label}: 
Can the annotator think of instances where the text can be considered as toxic or normal irrespective of which group makes the statement? This is most likely based on the annotator's value system and the sample in question need not contain any specific offensive words to be deemed offensive.

\textit{c. Inferred group label}: 
If the author of the text is a member of the inferred or target group (eg. muslims, blacks, women, etc) group, will the text be considered as toxic or normal? Sometimes a text may not be considered offensive by the audience if the author belongs to the same inferred group as the audience. Where it is inconceivable that a member of the inferred group would make such an utterance, the text should be classified as offensive.

\section{SBIC Opposition Rate}
\begin{table}[h]
  \caption{SBIC Opposition Rate}
  \label{data-summary_4}
  \centering
  \begin{tabular}{p{2cm}p{3cm}p{2cm}}
    \toprule
    Race &  No. of Samples \newline Annotated &   Opposition \newline Rate  \\
    \midrule
    White  & 117,616 \textit{(80\%)} & 12.2\% \\
    Hispanic    & 8,981 \textit{(6\%)}& 14.3\% \\
    Asian     & 8,719 \textit{(5\%)} & 14.6\% \\
    Black     & 5,522 \textit{(4\%)} & 17.4\% \\
    Native     & 550 \textit{(0.4\%} & 7.5\% \\
    Other     & 4,025 \textit{(3\%)} & 3\% \\
    NA     & 1,725 \textit{(1\%)} & 23.3\% \\
    \bottomrule
  \end{tabular}
\end{table}

\section{Sample Annotations for SBIC and Jigsaw Datsets}
\begin{table}[h]
  \caption{Sample Annotations for SBIC and Jigsaw Datsets}
  \label{data-summary_2}
  \centering
  \begin{tabular}{p{2cm}p{3cm}p{2cm}p{2cm}}
    \toprule
    No. of \newline Annotators &  Overall \newline Sample Size &   Offensive \newline Sample Size & Normal \newline Sample Size\\
    \midrule
    \textbf{SBIC}  \\
    1  & 1,539 \textit{(3.4\%)} & 1,268 & 271  \\
    2  & 7,540 \textit{(16.7\%)} & 6,371 & 1,169 \\
    3  & 27,275 \textit{(60.31\%)} & 8,851 & 18,424 \\
    4  & 4,008 \textit{(8.9\%)} & 4,001 & 7 \\
    5-10 & 4,807 \textit{(10.63\%)} & 4,798 & 9 \\
    11-15 & 50 \textit{(0.11\%)} & 50 & 0 \\
    16-20 & 4 \textit{(0.009\%)} & 4 & 0 \\
    \\
    \textbf{Jigsaw}  \\
    3  & 53 \textit{(0.003\%)} & 0 & 53  \\
    4  & 1,185,412 \textit{(65.68\%)} & 3,028 & 1,182,384 \\
    5-10  & 542,226 \textit{(30.04\%)} & 61,955 & 480,271 \\
    11-20  & 1,333 \textit{(0.07\%)} & 330 & 1,003 \\
    21-50 & 5,874 \textit{(0.33\%)} & 2,455 & 3,419 \\
    51-75 & 54,084 \textit{(3\%)} & 39,149 & 14,935 \\
    76-100 & 14,561 \textit{(0.81\%)} & 12,396 & 2,165 \\
    101-1000 & 96 \textit{(0.005\%)} & 85 & 11 \\
    Above 1000 & 1,235 \textit{(0.07\%)} & 686 & 549 \\
    \bottomrule
  \end{tabular}
\end{table}
\clearpage

\section{Inconsistent Annotation}
\begin{table}[h]
  \caption{Inconsistent Annotation}
  \label{data-summary_3}
  \centering
  \begin{tabular}{p{2cm}p{8cm}p{2cm}}
    \toprule
    Annotator ID &  Sample Text &  Label  \\
    \midrule
    \textbf{HateXplain} \\
    4 & \textit{"always a favorite scene kept the n***** in its place did not it"}& Offensive \\
    \\
    4 & \textit{"has stupid rich hoe"}  & Normal \\
    \\
    \textbf{SBIC} \\
    29 & \textit{"Where do you find a paralysed person? Where you left them."}  & Offensive \\
    \\
    29 & \textit{"Incels, what's your opinion on bisexual women?"} & Normal \\
    \\
    \bottomrule
  \end{tabular}
\end{table}

\section{Contextless Samples}
\begin{table}[h]
  \caption{Contextless Samples}
  \label{contextless-samples}
  \centering
  \begin{tabular}{p{2cm}p{8cm}}
    \toprule
    Dataset &  Sample Text \\
    \midrule
    HateXplain & \textit{"why Arabs lose wars"} \\
    \\
    SBIC & \textit{"good enough fag"} \\
    \\
    Jigsaw & \textit{"http://mustreadalaska.com/homer-city-council-goes-full-resist-trump-mode-resolution/"}\\
    \\
    \bottomrule
  \end{tabular}
\end{table}

\end{document}